\documentclass[conference]{IEEEtran}
\usepackage{caption}
\usepackage{algorithm}
\usepackage{algpseudocode}
\usepackage{graphicx}
\usepackage{subcaption}
\usepackage{bm}
\usepackage{mathtools}
\usepackage{amssymb}
\usepackage{multirow}
\usepackage{mathrsfs}
\usepackage{soul}
\usepackage{xcolor}
\usepackage{adjustbox}

\usepackage{enumitem}

\setenumerate[1]{itemsep=0.5pt,partopsep=0pt,parsep=\parskip, topsep=2pt, leftmargin=12pt,}

\newcommand*{\ToolName}{ZOBNN}

\begin{document}

\title{\huge{\ToolName: Zero-Overhead Dependable Design of Binary Neural Networks with Deliberately Quantized Parameters} }

\author{ Behnam Ghavami\textsuperscript{1}, Mohammad Shahidzadeh\textsuperscript{2}, Lesley Shannon\textsuperscript{2}, Steve Wilton\textsuperscript{1}\\
\textsuperscript{1}University of British Columbia, Canada\\ \textsuperscript{2}Simon~Fraser~University, Canada\\
\\
This paper has been accepted for publication at IOLTS 2024.
}
\IEEEoverridecommandlockouts        
\IEEEpubid{\makebox[\columnwidth]{ 979-8-3503-7055-3/24/\$31.00 \copyright2024 Crown \hfill} \hspace{\columnsep}\makebox[\columnwidth]{ }}
\maketitle
\IEEEpubidadjcol

\begin{abstract}
\small
Low-precision weights and activations in deep neural networks (DNNs) outperform their full-precision counterparts in terms of hardware efficiency.
When implemented with low-precision operations, specifically in the extreme case where network parameters are binarized (i.e. BNNs), the two most frequently mentioned benefits of quantization are reduced memory consumption and a faster inference process.
In this paper, we introduce a third advantage of very low-precision neural networks: improved fault-tolerance attribute. 
We investigate the impact of memory faults on state-of-the-art binary neural networks (BNNs) through comprehensive analysis. Despite the inclusion of floating-point parameters in BNN architectures to improve accuracy, our findings reveal that BNNs are highly sensitive to deviations in these parameters caused by memory faults.
In light of this crucial finding, we propose a technique to improve BNN dependability by restricting the range of float parameters through a novel deliberately uniform quantization. The introduced quantization technique results in a reduction in the proportion of floating-point parameters utilized in the BNN, without incurring any additional computational overheads during the inference stage. The extensive experimental fault simulation on the proposed BNN architecture (i.e. \ToolName) reveal a remarkable 5X enhancement in robustness compared to conventional floating-point DNN. Notably, this improvement is achieved without incurring any computational overhead. Crucially, this enhancement comes without computational overhead. \ToolName~excels in critical edge applications characterized by limited computational resources, prioritizing both dependability and real-time performance.

\end{abstract}


\section{Introduction}\label{sec:intro}

Deep neural networks (DNNs) have become increasingly important in a variety of application domains, including video analytics, autonomous vehicle systems, and medical devices, due to recent advances that improve their performance and energy efficiency. However, practical reliability considerations make the widespread real-world deployment of DNN-based systems difficult, particularly in many safety-critical domains \cite{10032608}. Recent research has shown that edge systems based on DNNs are susceptible to transient hardware faults resulting from particle strikes or voltage droops and can significantly impact the reliability of safety-critical DNN-based autonomous vehicles \cite{9586116}. For example, a single soft error may result in the misclassification of objects in the field of self-driving cars, which can cause the car to take inappropriate actions \cite{9806152}. This issue may worsen with the technology node scaling of hardware accelerators for autonomous machine computing. Therefore, there is a need for designing and implementing dependable DNN systems.


On the other hand, dependability in safety-critical real-time DNN-based embedded systems faces a unique challenge due to the requirements of low power, high throughput, and "low latency" where traditional fault-tolerant techniques, which include redundant-based hardware/software/data, can result in \textit{high overheads} in hardware cost, energy, and performance, making them impractical for real-time edge DNN systems. For instance, an autonomous vehicle must process driving data within a few milliseconds for real-world deployment and deploying traditional protection techniques may lead to delayed response times, potentially leading to reaction-time-based accidents. Given the resource constraints inherent in DNN-based edge applications and the critical need for real-time responsiveness, there exists a compelling imperative to develop a customized error mitigation technique. This prompts the central research question: \textit{Is it feasible to design highly fault-tolerant DNN models for safety-critical applications in embedded edge systems \textbf{without} incurring additional run-time and resource overhead?}

In this paper, we present the advantages of quantization as an enhanced cost-free fault-tolerance feature for DNNs. Quantization is defined as the reduction of precision used to represent neural network parameters, typically from $m$ bits to $n$ bits (where $m$ $<$ $n$). 
Our observation reveals that fault-free DNN neurons values typically fall within a limited range, a small fraction of the total range offered by data representation. Contrastingly, memory faults can lead to severe inference output corruptions when neurons values manifest significantly higher than the expected range. Based on this insight, \textit{we propose restricting the range of DNN parameters through quantization as a means of fault propagation prevention without incurring overheads}. Specifically, we endorse \textbf{binarization}, the extreme case of quantization, as a promising solution \textbf{to enhance DNN reliability}. However, the adoption of binarization presents a primary challenge in maintaining the high accuracy of a DNN. Addressing this concern, contemporary state-of-the-art (SOTA) Binary Neural Networks (BNNs) demonstrate reasonable accuracy via strategically incorporate some network components with floating-point data representation \cite{zhang2021fracbnn}, striking a delicate balance between the hardware efficiency gains of binarization and the accuracy demands of complex tasks. However, this inclusion of floating-point data representation could directly impact BNN reliability, as \textit{uncovered through our comprehensive fault characterization}.
Building upon this crucial observation, we propose a novel \textit{"selectively quantization"} technique to confine the range of these float parameters, thereby enhancing network dependability. Importantly, this proposed quantization technique introduces \textbf{no additional operations or modifications} to the network, mitigating the risk of extra computational burden. The experimental results emphasize a notable 5X improvement, on average, in reliability achieved by the proposed BNN model, named as \ToolName, compared to conventional DNNs. This advancement is showcased with up to 32.32\% and 24.81\% accuracy improvement on SOTA BNNs, Dorefanet\cite{zhou2016dorefa} and FracBNN\cite{zhang2021fracbnn}, at a fault rate of 1e-4. Impressively,  \ToolName~attains nearly identical accuracy compared to the baseline fault-free BNN models.

In summary, the main contributions of this paper are as follows:
\begin{itemize}
    \item Our investigation undertakes a thorough exploration into the implications of memory faults on DNNs, with a specific focus on illuminating the inherent dependability advantages of quantization. We meticulously elucidate the merits of binarization, the zenith of quantization, as an intricately designed advanced fault-tolerance feature tailored to fortify the resilience of DNNs.

    \item We introduce a novel quantization method poised to elevate the robustness of binary networks, specifically by directing attention to the overhead-aware quantization of floating-point parameters.
    \item The effectiveness of the proposed approach is substantiated through fault simulations conducted on SOTA BNNs.

\end{itemize}

\section{Related Work}
\label{sec:related}

Modifying the neural network architecture has been introduced to increase its error resilience, which can be done either during training or outside training. 
Dias et al. \cite{dias2010ftset} proposed a resilience optimization procedure that involves changing the architecture of the DNN to reduce the impact of faults. Schorn et al. \cite{schorn2019efficient} introduced a resilience enhancement technique that replicates critical layers and features to achieve a more homogeneous resilience distribution within the network. 
An error correction technique based on the complementary robust layer (redundancy mechanism) and activation function modification (removal mechanism) has been recently introduced in \cite{ali2020erdnn}. 
Hong et al. \cite{hong2019terminal} introduced an efficient approach to mitigate memory faults in DNNs by modifying their architectures using Tanh \cite{nwankpa2018activation} as the activation function. 
To mitigate errors, based on the analysis and the observations from \cite{liew2016bounded}, Hoang et al. \cite{hoang2020ft} presented a new version of the ReLU activation function to squash high-intensity activation values to zero, which is called Clip-Act. 
Ranger \cite{chen2020ranger} used value restriction in different DNN layers as a way to rectify the faulty outputs caused by transient faults. Recently, Ghavami et al. \cite{9774635} introduced a fine-grained activation function to improve the error resilience of DNNs.
Research (\cite{8536419}) emphasizes the critical need to detect errors in CNN execution to prevent their propagation to fully connected layers. They propose redesigning maxpool layers for improved error detection and potential correction. 
Libabno et al. (\cite{9319148}) conducted neutron beam experiments to assess the impact of quantization processes and data precision reduction on the failure rate of neural networks deployed on FPGAs. Their results show that adopting an 8-bit integer design increases fault-free executions by over six times compared to a 32-bit floating-point implementation, highlighting the advantages of lower precision formats in FPGA environments.
Investigating the reliability of CNNs, \cite{ruospo2021investigating} analyzed the effects of reduced bit widths and different data types (floating-point and fixed-point) on network parameters to find the optimal balance for CNN reliability.
Additionally, \cite{9843614} demonstrates that implementing fault-aware training significantly improves the resilience of Quantized Neural Networks to soft errors when deployed on FPGAs.
Finally, Cavagnero et al. (\cite{9897813}) introduce a zero-overhead fault-tolerant solution focusing on DNN redesign and retraining, which enhances DNN reliability against transient faults by up to one order of magnitude.

\section{Preliminaries and Definitions}
\subsection{Deep Neural Networks}
DNNs are a type of machine learning model that is designed to predict an output as accurately as possible given an input $X$. DNNs are composed of multiple layers, where each layer receives the output of the previous layer as its input. The input data is sequentially passed through each layer, with each layer performing a specific computation on the input and passing the result to the next layer.




The individual layers of a DNN $F_{i}$ are made up of neurons, which perform mathematical operations on the input data $X$. Each neuron is connected to the neurons in the previous layer $F_{i-1}$ via a set of weights $\theta_{i-1}$, which are determined during the training phase of the DNN. The weights determine the strength and direction of the connections between the neurons.
The computations performed by each layer of a DNN are typically non-linear, which allows the network to model complex relationships between the input and output data. The non-linearities in the layers are often introduced using activation functions, which transform the output of each neuron in a non-linear way.
The final layer of a DNN produces the output prediction for the input data. 
Equation \ref{eqn:dnn1} illustrates the DNN computation.
  \vspace{-0.2in}


 \begin{equation}
    \label{eqn:dnn1}
DNN_{\Theta}(\bm{X}) = (F_{N}^{\theta_{N}} \circ F_{N-1}^{\theta_{N-1}} \circ \dots \circ F_{1}^{\theta_{1}}) (\bm{X})
 \end{equation}

The DNN architecture is typically composed of a sequence of two types of layers, designed to capture the hierarchical structure of the input data:
\begin{itemize}
    \item The \textit{first-type} consists of convolution and linear layers. Each neuron in these layers $N^{i}$ is computed as a sum of the product of the neurons in the previous layer $N^{i-1}$ to their corresponding weights $W^{k}$.
    \[ N^{(i,k)} =\sum_{j}^{} (W^{(k,j)})(N^{(i-1,j)}) \]
    \item  The \textit{second-type} of layer which consists of batchnormalization layers, activation functions, and others, were developed to increase the accuracy of DNNs. In DNNs, there is always at least one of the "second-type" layers in between the two "first-type" layers.
\end{itemize}

Therefore, the comprehensive DNN architecture can be visualized as below, with 'S' denoting a second-type and 'L' indicating a first-type layer:

\vspace{-0.1in}
 \begin{equation}
    \label{eqn:dnn2}
DNN_{\Theta}(\bm{X}) = ({L}_{N}^{\theta_{N}} \circ {S}_{N-1}^{\theta_{N-1}} \circ {L}_{N-2}^{\theta_{N-2}} \circ {S}_{N-3}^{\theta_{N-3}} \circ \dots {S}_{2}^{\theta_{2}} \circ {L}_{1}^{\theta_{1}}) (\bm{X})
 \end{equation}
\vspace{-0.3in}



\subsection{Integer Only Quantization of Neural Networks}

Integer-only quantization \cite{jacob2018quantization} plays a pivotal role in deep learning, especially for deploying models on resource-constrained hardware platforms. It involves reducing numerical precision to integers, allowing for efficient implementation of neural network operations using only integer arithmetic. This technique relies on an affine mapping of integers \( q \) to real numbers \( r \), expressed as \( r = \Delta(q - Z) \), where \( \Delta \) represents the scale factor and \( Z \) denotes the zero-point.

Utilizing this quantization scheme, the quantized representation of a neuron \( k \) in the \( i \)th layer can be expressed as \( N^{(i,k)} = \Delta_3(q_{3}^{(i,k)} - Z_3) \), while that of neuron \( j \) in the \( (i-1) \)th layer is \( N^{(i-1,j)} = \Delta_2(q_{2}^{(i-1,j)} - Z_2) \), reflecting varying scaling factors across layers. Furthermore, the \( j \)th weight of neuron \( k \) can be quantized as \( W^{(k,j)} = \Delta_3(q_{1}^{(k,j)} - Z_1) \). Employing this technique, neuron computation proceeds as follows:

\[ q^{(i,k)}_3 = Z_3 + M \sum_{j=1}^{N} (q^{(k,j)}_1 - Z_1)(q^{(i-1,j)}_2 - Z_2) \]

where \( M \), defined as \( M := \frac{\Delta_1 \Delta_2}{\Delta_3} \), first dequantizes the results from the previous layer by scaling them with \( \Delta_1 \times \Delta_2 \), then quantizes them again by applying \( \Delta_3 \).





\subsection{Binary Neural Networks}
\label{sec:BNN}

BNNs are the extreme case of quantization where the bit-width is one. In BNN, the intermediate L-type layer's parameters coupled with their inputs are passed through a \textit{binarization function} B. This allows the BNNs to run on more efficient hardware with XNOR and Pop-Count operations. While BNNs offer several advantages over DNNs, including a low memory footprint and power efficiency due to the use of binary values for weights and activations, they also suffer from a major disadvantage: \textit {a significant loss in accuracy that occurs after the binarization process.}
To mitigate this loss, existing BNN models\cite{zhou2016dorefa}\cite{zhang2021fracbnn} often use floating-point parameters in the first and the last L-type layers and in all S-type layers as described in equation \ref{eqn:dnn3}.
 \vspace{-0.1in}
 \begin{equation}
    \label{eqn:dnn3}
\begin{split}
BNN_{\Theta}(\bm{X}) = ({L}_{N}^{\theta_{N}} \circ {S}_{N-1}^{\theta_{N-1}} \circ {L}_{N-2}^{\textbf{B}(\theta_{N-2})} \circ \textbf{B} \circ {S}_{N-3}^{\theta_{N-3}}
\\ \circ \dots \circ {S}_{2}^{\theta_{2}} \circ {L}_{1}^{\theta_{1}}) (\bm{X})
\end{split}
\end{equation}

By using a mix of binary and floating-point representations, they can strike a balance between the memory efficiency of binary values and the higher precision of floating-point values, while still providing significant improvements in resource efficiency over traditional DNNs.

 \section{Motivation: Reliability analysis of Conventional BNNs}\label{sec:motivation}
\begin{figure}[!h]
 	\begin{center}
 	    \vspace{-0.1in}
 		\includegraphics[width = 0.5\textwidth]{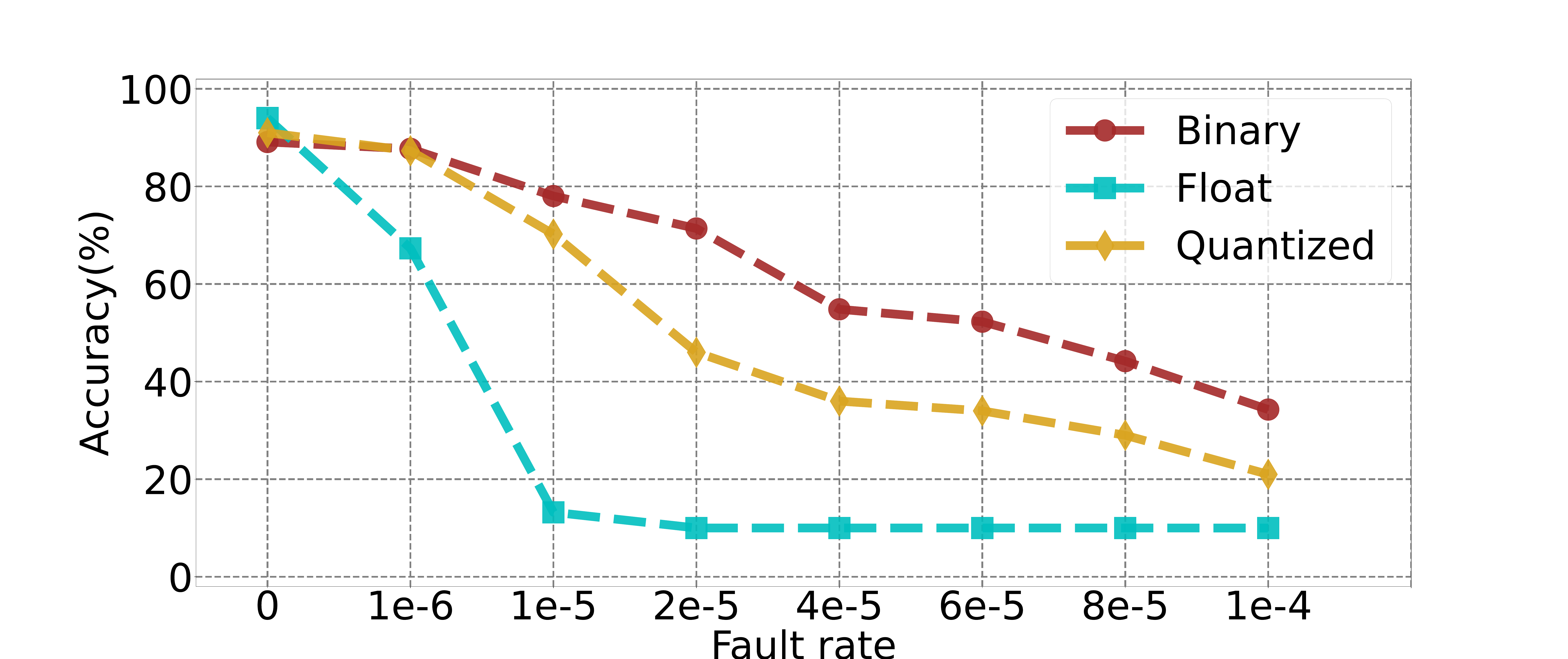}
 		\vspace{-0.2in}
 		\caption{Comparing accuracy drop points for fault rates in float, quantized, and binary models using the fracbnn\cite{zhang2021fracbnn} base architecture which further transformed to binary and quantized.}
 		\vspace{-0.1in}
 		\label{fig:BFCompare}
 	\end{center}	
 \vspace{-0.2in}
 \end{figure}

Despite the numerous advantages offered by BNNs, their resilience to memory faults has received little attention. 
As a result, there is a pressing need to better understand the vulnerability of BNNs and their effective safeguards to these types of faults. Figure \ref{fig:BFCompare} illustrates the impact of memory bit-flip faults on a floating-point DNN, its corresponding quantized version, and a BNN, at varying memory bit-flip probabilities (fault rates). In the DNN scenario, the accuracy drops in the 1e-6 fault rate and becomes a random output in the 1e-4 value. However, in the BNN case, the accuracy remains nearly the same in 1e-5 and gets a 50\% deviation from the baseline accuracy in the extreme case of 1e-4 fault rate. Our observation reveals fault-free neurons usually have values within a narrow range, significantly smaller than the total range in floating-point representation. We can state that the memory faults will propagate to the DNN output only if the affected neuron values exceed the expected range, leading to substantial output changes.
Consequently, \textit{"quantization" can then limit computation parameter ranges, serving as a \textbf{built-in mechanism to "fault propagation prevention"}}.

\begin{figure}[!h]
 	\begin{center}
 	    \vspace{-0.2in}
 		\includegraphics[width = 0.5\textwidth]{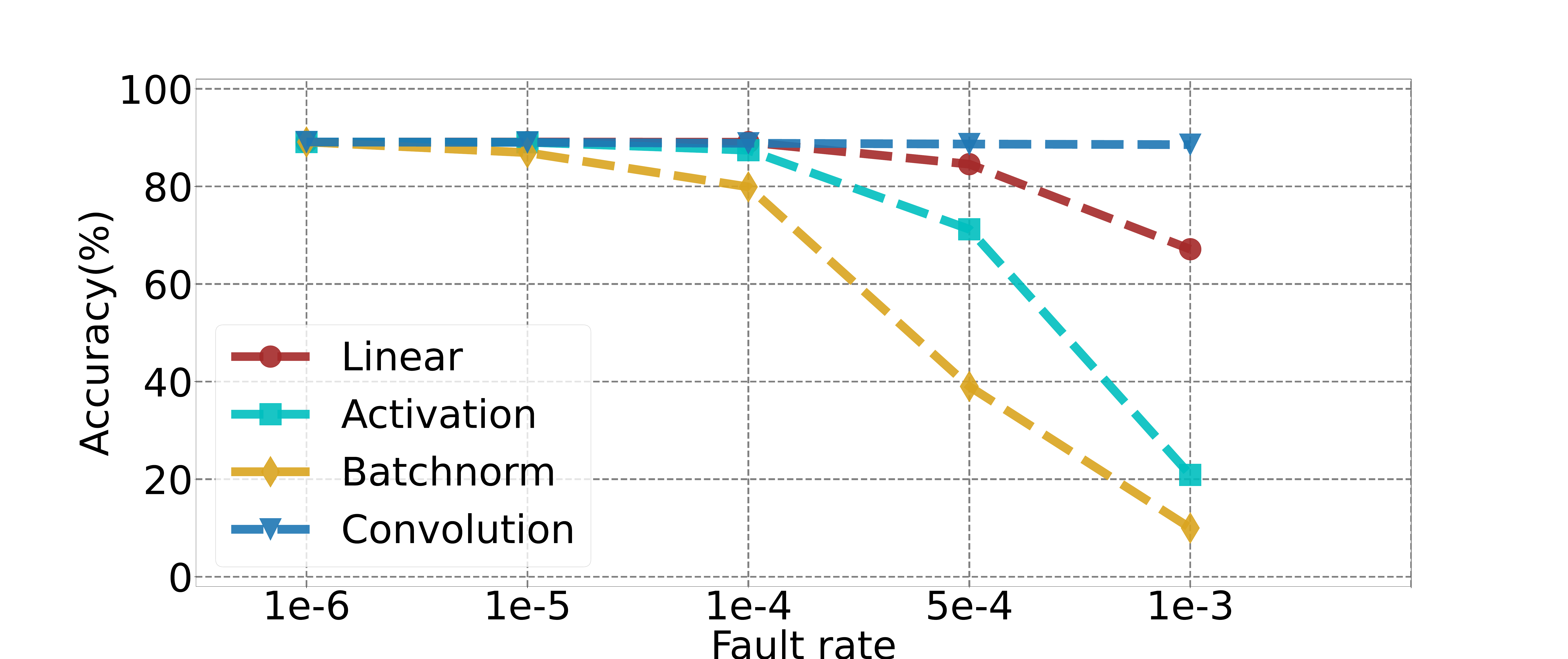}
 		\vspace{-0.2in}
        \caption{A comparison between different layers' accuracy drop-points under various fault rates in the fracbnn\cite{zhang2021fracbnn} architecture.}
 		\vspace{-0.1in}
 		\label{fig:LayerCompair}
 	\end{center}	
 \vspace{-0.2in}
 \end{figure}

Additionally, to comprehend the \textit{underlying vulnerability in BNN due to memory faults}, we conducted an in-depth fault analysis of individual layers on a BNN \cite{zhang2021fracbnn}. Figure \ref{fig:LayerCompair} illustrates these comparative results. Among the layers, L-type layers which are convolution layers exhibit the least significant accuracy drop, since all the parameters in the convolution layers are binarized as described in Section \ref{sec:BNN}. However, this is not the case for the remaining S-type layers and the final L-type layer, as their parameters are stored in floating-point format. \textit{The remaining layers' accuracy starts deviating from the baseline in a lower fault rate}. This aligns with the observation that soft errors can result in substantial corruption of inference outputs in floating-point's DNNs, especially when these errors manifest as values exceeding the expected range. Binary values, constrained to a limited range, and bit-flips have no opportunity to manifest as large value deviation. It is noteworthy that even after binarizing all Convolution layers, their memory footprint still dominates compared to other layers while exhibiting the smallest drop.
 \vspace{-0.1in}



\section{Proposed Selective Quantization of BNNs}
\label{sec:proposed}

To enhance the reliability of BNNs while addressing computation overhead, we introduce a novel selctive quantization method for the S-type layers, integrated with the first and last L-type layers within a BNN. This preserves baseline accuracy without adding extra computational overhead. 

Equation \ref{eqn:dnn4} exemplifies a conventional quantization approach involves three components for each layer. Firstly, a quantization function $Q(\theta_{N})$ is applied to the parameters, facilitating the transition from floating-point to integer representation. Subsequently, an additional quantization layer $Q$ is incorporated into the input, enabling the entire layer to operate seamlessly in quantized computation form. Lastly, a dequantization layer $D$ restores the quantized result to its original form.
\vspace{-0.15in}

 \begin{equation}
    \label{eqn:dnn4}
\begin{split}
DNN_{\Theta}(\bm{X}) = (D \circ {L}_{N}^{Q(\theta_{N})} \circ Q \circ D \circ {S}_{N-1}^{Q(\theta_{N-1})} \circ Q \circ\\ {L}_{N-2}^{B(\theta_{N-2})} \circ B \circ \dots \circ D \circ {S}_{2}^{Q(\theta_{2})} \circ Q \circ D \circ {L}_{1}^{Q(\theta_{1})} \circ Q) (\bm{X})
\end{split}
 \end{equation}

However, this approaches introduce supplementary extra parameters and computational load. Additionally, storing the parameters of the quantization and dequantization layers in memory poses a \textit{potential source of failure}, impacting the reliability of the entire network. To alleviate these concerns and eliminate the need for the quantization (Q) and dequantization (D) layers throughout the network, several modifications are proposed. Primarily, we established a \textbf{reciprocal relationship between Q and D}, ensuring \(Q(D(x)) = x\). This adjustment allows for the removal of consecutive Q and D functions, as demonstrated in Equation \ref{eqn:dnn5}.
\vspace{-0.15in}

 \begin{equation}
    \label{eqn:dnn5}
\begin{split}
DNN_{\Theta}(\bm{X}) = (D \circ {L}_{N}^{Q(\theta_{N})} \circ {S}_{N-1}^{Q(\theta_{N-1})} \circ Q \circ\\ {L}_{N-2}^{B(\theta_{N-2})} \circ B \circ \dots \circ {S}_{2}^{Q(\theta_{2})} \circ {L}_{1}^{Q(\theta_{1})} \circ Q) (\bm{X})
\end{split}
 \end{equation}

Executing this operation allows for the elimination of the majority of quantization and dequantization functions. Nonetheless, certain corner cases persist. Specifically, some S-type layers receive input after L-type layers, introducing a single Q-type layer before them. For instance, layer \(S_{N-1}\) exemplifies such a scenario. To eliminate the quantization function in these layers, it is imperative that every S-type layer can seamlessly process both quantized and dequantized inputs. Subsequent Q and D functions are linked with the input and output sections, which we would discuss separately in Sections \ref{seq:Output} and \ref{seq:Input}. Finally, Section \ref{seq:Binary} delves into the \(B \circ D\) components and the methods employed to discard them. These operations culminate in the final BNN represented by Equation \ref{eqn:dnn6}, devoid of any additional parameters that might compromise its robustness.
\vspace{-0.15in}

 \begin{equation}
    \label{eqn:dnn6}
\begin{split}
DNN_{\Theta}(\bm{X}) = ({L}_{N}^{Q(\theta_{N})} \circ {S}_{N-1}^{Q(\theta_{N-1})} \circ\\ {L}_{N-2}^{B(\theta_{N-2})} \circ B \circ \dots \circ {S}_{2}^{Q(\theta_{2})} \circ {L}_{1}^{Q(\theta_{1})} ) (\bm{X})
\end{split}
 \end{equation}

 

\subsection{Dequantization and Quantization Layers}
\label{seq:dequan}

The following equation illustrates the proposed quantization and dequantization layers of \ToolName. The quantization layer transforms the real floating-point number $x_{r}$ into its quantized integer counterpart $x_{q}$ using the parameter $\Delta$. The parameter $\Delta$ is determined by the equation $\frac{max(\theta)}{2^{\#bits-1}-1}$. This signifies that we utilize the maximum integer, $2^{\#bits-1}-1$, to represent the largest parameter in the model, denoted as $max(\theta)$. Consequently, the parameters are quantized within the range of $-max(\theta)$ to $max(\theta)$.


\begin{equation}
    \label{eqn:quan1}
Q(x_{r}) = round(x_{r}/\Delta) = x_{q} 
\end{equation}

Next, the dequantization function in Equation \ref{eqn:quan3} should transfer the quantized number to integer format. Also, it should be such that Q(D(x)) = x. Equation \ref{eqn:quan4} illustrates this reciprocal relation between Q and D.
\vspace{-0.1in}
\begin{equation}
    \label{eqn:quan3}
D(x_{q}) \approx \Delta \times (x_{q}) = x_{r}
\end{equation}
\vspace{-0.20in}
\begin{equation}
    \label{eqn:quan4}
Q(D(x_{q})) \approx round(\frac{\Delta \times (x_{q})}{\Delta}) = x_{q}
\end{equation}

\subsection{Input Layer}
\label{seq:Input}
Following the application of common quantization to the input layer, the input is processed through the Q layer, denoted as ${L}{1}^{Q(\theta{1})} \circ Q \circ x_{r}$. By omitting the Q layer in this equation, the layer receives its input directly in the $x_r$ format. The subsequent equation elucidates the quantization method employed for this layer.
\vspace{-0.20in}

\begin{equation}
\begin{gathered}
\vspace{-0.20in}
conv(X) = X*W = X_{r}*W_{q} \times \Delta =  \Delta \times \underbrace{X_{r}*W_{q}}_\text{$\gamma$}\\
 conv(\Delta \times X_{q}) =\Delta \times \gamma\\
 conv(X_{q}) =\gamma  
\end{gathered}
\end{equation}


\subsection{Output Layer}
\label{seq:Output}
The output layer in the majority of BNNs typically remains unbinarized. As a result, this layer will undergo standard quantization. In the standard quantization, the output layer is succeeded by a D-type layer. The removal of this D-type layer proves to be straightforward in neural networks dedicated to detection tasks, as it is succeeded by an argmax function. Notably, \( \text{Argmax}(D(x)) = \text{Argmax}\left(\frac{x}{\Delta}\right) = \text{Argmax}(x) \), thereby automatically eliminating the necessity for the dequantization function.




\subsection{Binarized Layer}
\label{seq:Binary}
The sign layer B serves as the antecedent function for the binarized layers in the network. Following standard quantization, the input to this function is derived from a D layer, denoted as $B \circ D$. If the dequantization layer is excluded, the input to the sign layer becomes $x_{q}$. The subsequent equation depicts this layer when receiving an input of $x_{q}$. The sole modification required for this function involves dividing the original constants by $\Delta$, yielding identical results for an input of $x_{r}$.
\vspace{-0.1in}

\begin{equation}
\begin{gathered}
 sign(X) = \frac{clamp(X,1,-1)+1}{delta}\\ 
 sign(\Delta \times X) = \frac{clamp(X_{q}\Delta,1,-1)+1}{delta}\\
 sign(X_{q}) = \underbrace{ \frac{clamp(X_{q},\frac{1}{\Delta},-\frac{1}{\Delta})+\frac{1}{\Delta}}{\frac{delta}{\Delta}} }_\text{$\gamma$}\\
 sign(X_{q})=\gamma
\end{gathered}
\end{equation}

\subsection{Other Layers via SOTA BNN}
All the S-type layers should work with both $x_{q}$ and $x_{r}$ inputs and give their output in $x_{q}$ format. Here we show the quantization of S-type layers for two common layers, batch-norm and rprelu, in SOTA BNNs.




\subsubsection{batch-norm}
The batch normalization layer is a ubiquitous component in BNN architecture, comprising four parameters available for quantization. When dealing with quantized input, all four parameters are quantized, whereas in the case of real input, only W and B undergo quantization. This approach ensures a consistently quantized output regardless of the input nature. Equation \ref{eqn:batch1} illustrates the quantization of this function for an input $x_q$, and equation \ref{eqn:batch2} does so for an input $x_r$.

\begin{equation}
\label{eqn:batch1}
\begin{gathered}
batchnorm(X) = W \times \frac{X_{r}-\mu}{\sigma} + B  \\
= \Delta \times W_{q} \times  \frac{\Delta \times X_{q}-\Delta \times \mu_{q}}{\Delta \times \sigma_{q}}+\Delta \times B_{q} \\
= \Delta \times \underbrace{W_{q} \times  \frac{X_{q}-\mu_{q}}{\sigma_{q}}+B_{q}}_\text{$\gamma$} \\
bathcnomr(\Delta \times X_{q}) =\Delta \times \gamma \\
batchnorm(X_{q}) =\gamma 
\end{gathered}
\end{equation}

\begin{equation}
\label{eqn:batch2}
\begin{gathered}
batchnorm(X) = W \times \frac{X_{r}-\mu}{\sigma} + B  \\
= \Delta \times W_{q} \times  \frac{X_{r}- \mu_{r}}{ \sigma_{r}}+\Delta \times B_{q} \\
= \Delta \times \underbrace{W_{q} \times  \frac{X_{r}-\mu_{r}}{\sigma_{r}}+B_{q}}_\text{$\gamma$} \\
bathcnomr(X_{r}) =\Delta \times \gamma
\end{gathered}
\end{equation}


\subsubsection{rprelu}
The rprelu layer functions as the activation function in SOTA BNNs, comprising three parameters, two of which are biases, and a parameter W which equals 1 when $x>-B1$ and is constant otherwise. The following two equations illustrate the quantization of these functions given inputs of $x_q$ and $x_r$. In the first scenario, the parameters $B1$ and $B2$ are quantized, while in the second scenario, it's the $W$ and $B2$. This quantization ensures a quantized output in both cases.
\begin{equation}
\begin{gathered}
 rprelu(X) = W \times (X+B1) + B2 \\
 =
       W_{r} \times (\Delta \times X_{q}+ \Delta \times B1_{q}) + \Delta \times B2_{q}\\
= \Delta \times \underbrace{W_{r} \times ( X_{q}+ \times B1_{q}) + \times B2_{q}}_\text{$\gamma$}\\
 rprelu(\Delta \times X_{q}) =\Delta \times \gamma\\
 rprelu(X_{q}) =\gamma  
\end{gathered}
\end{equation}

\begin{equation}
\begin{gathered}
 rprelu(X) = W \times (X+B1) + B2 \\
 = \Delta \times  W_{r} \times (X_{r}+ B1_{r}) + \Delta \times B2_{q}\\
 = \Delta \times \underbrace{W_{q} \times ( X_{r}+ \times B1_{r}) + \times B2_{q}}_\text{$\gamma$}\\
 rprelu(X_{r}) =\Delta \times \gamma
\end{gathered}
\end{equation}
\section{Experimental Fault Simulation Setup}
\label{sec:results}

\subsection{Fault Injection Framework}
To evaluate the error tolerance of a DNN, intentional simulated faults are introduced at various levels of the abstraction. Employing software-level fault simulation proves to be the optimal approach due to its speed and reasonably accurate results, especially considering the typically prolonged simulation time for large networks. In this study, a dynamic software-level fault injection tool on top of PyTorch framework is developed throughout the BNN inference process. Fault models are implemented as native tensor operations to minimize any potential performance impact. 
This involves flipping every bit in the memory with a probability denoted as P, referred to as the \textit{"fault rate"} in our investigation.
Subsequently, 
to ensure robustness, this fault study is repeated 500 times, and the average across all iterations is calculated.

\subsection{Fault Rate Value}
To maintain the generality of our investigation across different memory hardware and environmental setups, we analyze the impact of faults within the fault rate range of 1e-6 to 1e-3, aligning with recent experiments. Reagen et al. \cite{reagen2016minerva} observed SRAM fault rates of approximately 1e-5 at a voltage reduction to 0.7 volts. Nguyen et al. \cite{nguyen2019st} delved into DRAM intricacies, revealing fault rate fluctuations from 0 to 2e-3 at a 1s refresh rate within the temperature range of $25^{o}C$ to $90^{o}C$. They also demonstrated fault rate variations from 1e-7 to 2e-3 at $60^{o}C$ by adjusting the refresh rate from 1s to 10s. 





\subsection{BNN Models, Dataset and Inference Setup}

The fault study was carried out within FracBNN\cite{zhang2021fracbnn} and Dorefanet\cite{zhou2016dorefa}-—the state-of-the-art BNNs. Our investigation focused on two well-established image classification datasets, CIFAR-10 and ImageNet. FracBNN, based on the ResNet architecture, demonstrates a baseline accuracy of 89.1\% on CIFAR-10 and 71.7\% on the ImageNet dataset, while Dorefanet achieves 93\% accuracy on the CIFAR-10 dataset.

In measuring the inference time, BMXNet\cite{yang2017bmxnet} was employed on an Intel Core i7-5500U CPU. The networks operated in inference mode on the CIFAR-10 test dataset with a batch size of 1 and we recorded the average time over 20 tests.



\section{Experimental Results}
\label{sec:results}
\subsection{Dependability Investigation}
\begin{figure}[!h]
 	\begin{center}
 	    \vspace{-0.25in}
 		\includegraphics[width = 0.5\textwidth]{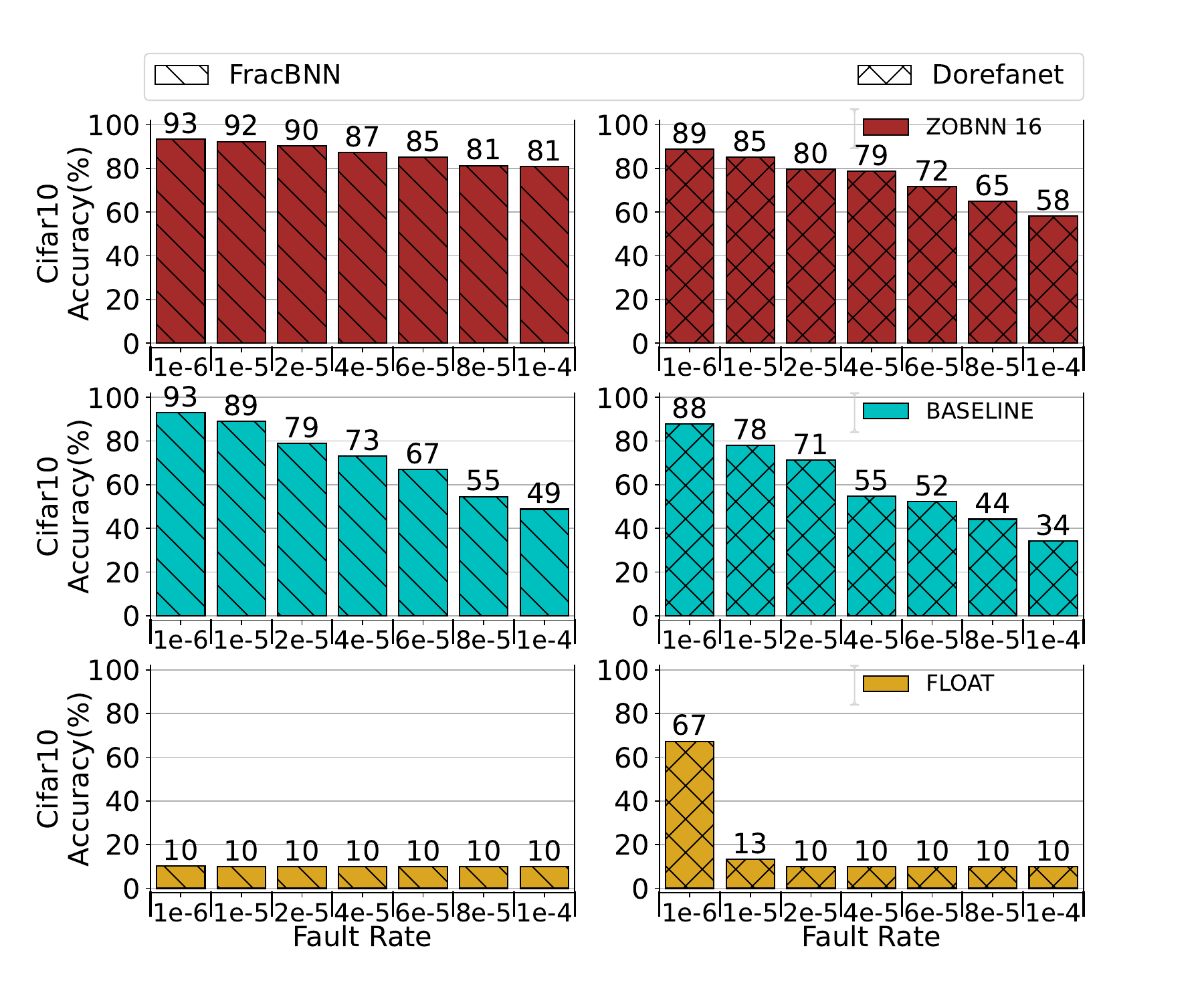}
 		\vspace{-0.4in}
 	\caption{Comprehensive fault study of float, baseline and proposed \ToolName{}-based FracBNN \cite{zhang2021fracbnn} and DoReFaNet \cite{zhou2016dorefa}.}
 		\vspace{-0.1in}
 		\label{fig:Final_resul}
 	\end{center}	
 \vspace{-0.1in}
 \end{figure}

Figure \ref{fig:Final_resul} presents a dependability comparison among the the baseline BNN, an equivalent floating-point network matched with the binary counterpart, and proposed 16-bit quantized \ToolName{} under various fault rates. In the case of DorefaNet, \ToolName{} achieves remarkable dependability results, maintaining accuracy above 78\% even with a fault rate of 1e-4, while the baseline BNN accuracy drops below 48\%. For FracBNN, the baseline BNN accuracy reaches 54\% at the 4e-5 fault rate and declines to 34\% at the 1e-4 fault rate.
However, \ToolName{} attains 78\% accuracy at a fault rate of 4e-5 and exceeds 58\% in the extreme case of a 1e-4 fault rate. Notably, the floating-point counterpart consistently generates random outputs in all scenarios. This highlights \ToolName{} as presenting \textbf{a remarkable 5X enhancement in robustness}, on average, compared to conventional floating-point DNNs.

\begin{figure}[!h]
 	\begin{center}
 	    \vspace{-0.1in}
 		\includegraphics[width = 0.4\textwidth]{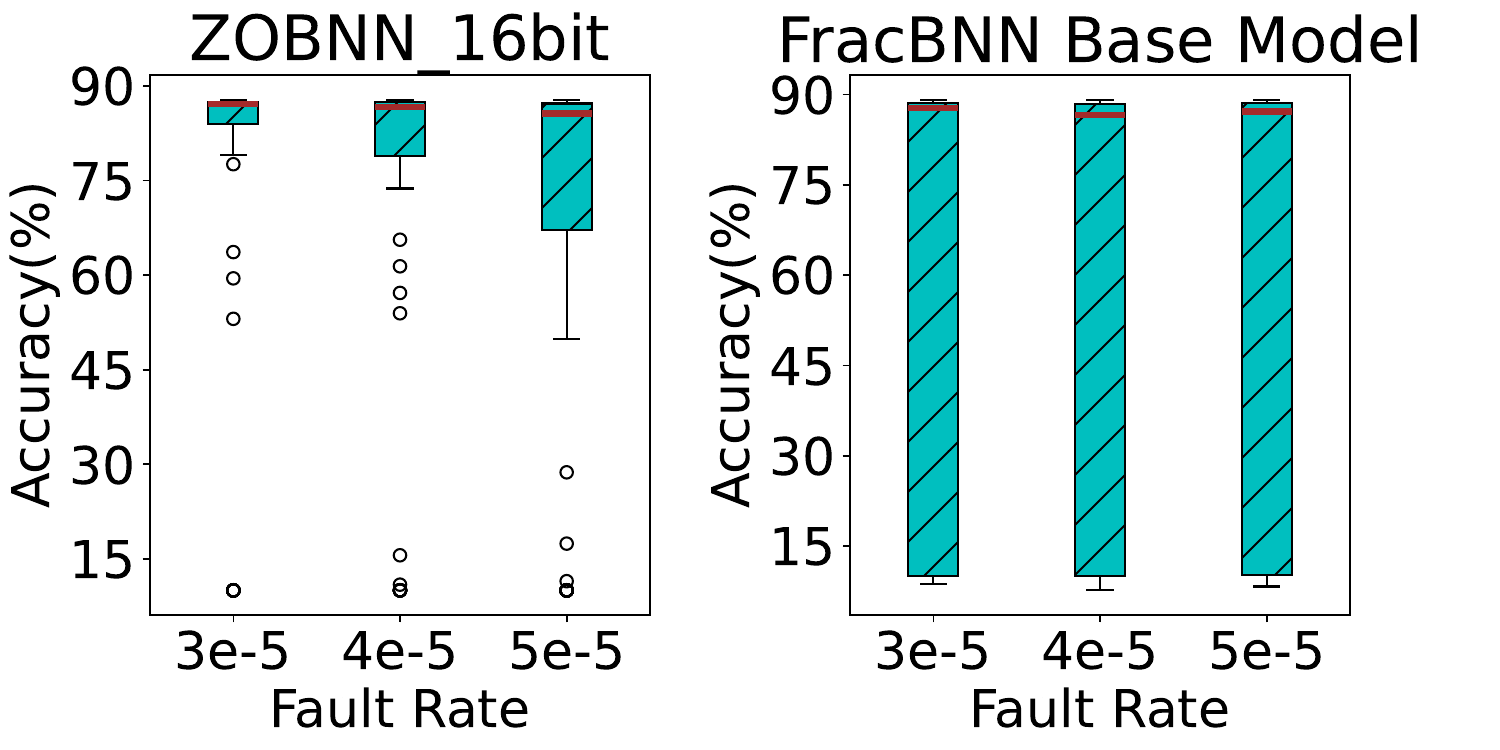}
 		\vspace{-0.1in}
    \caption{Comparison of accuracy distribution in FracBNN\ \cite{zhang2021fracbnn} model to its selective quantized model under \ToolName{}. The distribution was drawn from 500 different samples under three different fault rates.}
 		\vspace{-0.1in}
 		\label{fig:distribution}
 	\end{center}	
 \vspace{-0.01in}
 \end{figure}

Figure \ref{fig:distribution} depicts a comparative analysis of the accuracy distribution between the base Fracbnn and the proposed \ToolName{}. While the median remains consistent in both instances, our approach exhibits a significantly narrower range of error. This tightens the fault margin, resulting in a notably higher average accuracy.

\subsection{Runtime and Memory Overhead}
Table \ref{Tab:ac_md} illustrates the impact of \ToolName{} on the memory footprint of FracBNN. \ToolName{} enables the network to achieve quantization up to 8 bits in CIFAR-10 mode, incurring a 20.1\% memory reduction. Furthermore, \ToolName{} achieves 12-bit quantization on the ImageNet dataset, resulting in a 34\% memory reduction.

Table \ref{Tab:time} presents the runtime overhead of FracBNN and DorefaNet on the CIFAR-10 dataset. Notably, \ToolName{} introduces \textit{no additional computational overhead} to the network, in contrast to conventional quantization method, which incurs a 26.6\% and 22.65\% computational overhead on FracBNN and DorefaNet, respectively.

\begin{table}[tb!]
\begin{center}
\caption{
Analyzing the impact of performing \ToolName{} selective quantization on FracBNN's baseline accuracy and memory footprint of the model} across ImageNet and CIFAR-10 datasets.
\vspace{-0.05in}    
\scalebox{0.90}{
\begin{tabular}
{|| c | c | c | c | c | c | c ||}
\hline
\textbf{Dataset}&\textbf{AC/MD}&\multicolumn{5}{c|}{\textbf{Number of quantization bits}}\\
\cline{3-7}
&&16&14&12&10&8\\
\hline \hline
ImageNet&Baseline accuracy& 71.5\% & 71.5\%  & 71.3\% & 66.5\% & 0.1\%\\
\cline{2-7}
&Memory reduction& 27.2\% & 30.6\%  & 34\% & 37.4\% & 40.8\%\\
\hline
Cifar10&Baseline accuracy& 88.9\% & 88.5\%  & 88.8\% & 88.4\% & 87.1\%\\
\cline{2-7}
&Memory reduction& 13.4\% & 15.1\%  & 16.7\% & 18.4\% & 20.1\%\\
\hline
\end{tabular}
}
   \label{Tab:ac_md}
\end{center}
\vspace{-0.1in}
\end{table}



\begin{table}[tb!]
\begin{center}
\vspace{0.01in}
\caption{Comparison of inference times (in milliseconds) for floating-point DNNs, 16-bit conventional quantization, and 16-bit quantized \ToolName{} on the CIFAR-10 test-set.}

\vspace{-0.05in}    
\scalebox{0.96}{
\begin{tabular}{|| c | c | c | c ||}
\hline
&\multicolumn{3}{c|}{\textbf{Time in ms for 3 different binary network}}\\
\hline
\textbf{Architecture}&\textbf{Float DNN}&\textbf{Conventional quantization}&\textbf{\ToolName{}}\\ 
\hline \hline
\textbf{$FracBNN$ \cite{zhang2021fracbnn}}&3.46&4.72&3.45\\
\hline
\textbf{$Dorefanet$ \cite{zhou2016dorefa}}&0.97&1.26&0.95\\
\hline
\end{tabular}
}
   \label{Tab:time}
\end{center}
\vspace{-0.2in}
\end{table}


\section{Conclusion}
\label{sec:concl}

In the dynamic realm of edge deep learning applications, reliability is paramount. This paper delves into fault injection experiments, unveiling the inherent robustness of BNNs against memory faults. A novel deliberately quantization method takes center stage, fortifying BNN reliability without burdensome computational costs. Experimental results highlight the transformative power of proposed \ToolName{} as it significantly elevates BNN reliability while simultaneously streamlining memory footprint. This breakthrough opens doors for DNN deployment in the harshest embedded environments, where fault rates run high and resources are constrained.

\section{Acknowledgements}
We acknowledge the support from the Natural Sciences and Engineering Research Council (NSERC) of Canada, This work is funded by the Natural Sciences and Engineering Research Council of Canada nserc https://www.nserccrsng.gc.ca/ under Grant No. NETGP485577-15 nserc (COHESA project) and 341516 NSERC (RGPIN), along with in-kind support from AMD and Intel/altera.

\bibliographystyle{IEEEtran}
\small
\bibliography{references}	

\begin{thebibliography}{10}
\providecommand{\url}[1]{#1}
\csname url@samestyle\endcsname
\providecommand{\newblock}{\relax}
\providecommand{\bibinfo}[2]{#2}
\providecommand{\BIBentrySTDinterwordspacing}{\spaceskip=0pt\relax}
\providecommand{\BIBentryALTinterwordstretchfactor}{4}
\providecommand{\BIBentryALTinterwordspacing}{\spaceskip=\fontdimen2\font plus
\BIBentryALTinterwordstretchfactor\fontdimen3\font minus \fontdimen4\font\relax}
\providecommand{\BIBforeignlanguage}[2]{{%
\expandafter\ifx\csname l@#1\endcsname\relax
\typeout{** WARNING: IEEEtran.bst: No hyphenation pattern has been}%
\typeout{** loaded for the language `#1'. Using the pattern for}%
\typeout{** the default language instead.}%
\else
\language=\csname l@#1\endcsname
\fi
#2}}
\providecommand{\BIBdecl}{\relax}
\BIBdecl

\bibitem{10032608}
F.~Su, C.~Liu, and H.-G. Stratigopoulos, ``Testability and dependability of ai hardware: Survey, trends, challenges, and perspectives,'' \emph{IEEE Design \& Test}, vol.~40, no.~2, pp. 8--58, 2023.

\bibitem{9586116}
Z.~Wan, A.~Anwar, Y.-S. Hsiao, T.~Jia, V.~J. Reddi, and A.~Raychowdhury, ``Analyzing and improving fault tolerance of learning-based navigation systems,'' in \emph{2021 58th ACM/IEEE Design Automation Conference (DAC)}.\hskip 1em plus 0.5em minus 0.4em\relax IEEE, 2021, pp. 841--846.

\bibitem{9806152}
B.~Ghavami, S.~Movi, Z.~Fang, and L.~Shannon, ``Stealthy attack on algorithmic-protected dnns via smart bit flipping,'' in \emph{2022 23rd International Symposium on Quality Electronic Design (ISQED)}.\hskip 1em plus 0.5em minus 0.4em\relax IEEE, 2022, pp. 1--7.

\bibitem{zhang2021fracbnn}
Y.~Zhang, J.~Pan, X.~Liu, H.~Chen, D.~Chen, and Z.~Zhang, ``Fracbnn: Accurate and fpga-efficient binary neural networks with fractional activations,'' in \emph{The 2021 ACM/SIGDA International Symposium on Field-Programmable Gate Arrays}, 2021, pp. 171--182.

\bibitem{zhou2016dorefa}
S.~Zhou, Y.~Wu, Z.~Ni, X.~Zhou, H.~Wen, and Y.~Zou, ``Dorefa-net: Training low bitwidth convolutional neural networks with low bitwidth gradients,'' \emph{arXiv preprint arXiv:1606.06160}, 2016.

\bibitem{dias2010ftset}
F.~M. Dias, R.~Borralho, P.~Fontes, and A.~Antunes, ``Ftset-a software tool for fault tolerance evaluation and improvement,'' \emph{Neural Computing and Applications}, vol.~19, pp. 701--712, 2010.

\bibitem{schorn2019efficient}
C.~Schorn, A.~Guntoro, and G.~Ascheid, ``An efficient bit-flip resilience optimization method for deep neural networks,'' in \emph{2019 Design, Automation \& Test in Europe Conference \& Exhibition (DATE)}.\hskip 1em plus 0.5em minus 0.4em\relax IEEE, 2019, pp. 1507--1512.

\bibitem{ali2020erdnn}
M.~S. Ali, T.~B. Iqbal, K.-H. Lee, A.~Muqeet, S.~Lee, L.~Kim, and S.-H. Bae, ``Erdnn: Error-resilient deep neural networks with a new error correction layer and piece-wise rectified linear unit,'' \emph{IEEE Access}, vol.~8, pp. 158\,702--158\,711, 2020.

\bibitem{hong2019terminal}
S.~Hong, P.~Frigo, Y.~Kaya, C.~Giuffrida, and T.~Dumitraș, ``Terminal brain damage: Exposing the graceless degradation in deep neural networks under hardware fault attacks,'' in \emph{28th USENIX Security Symposium (USENIX Security 19)}, 2019, pp. 497--514.

\bibitem{nwankpa2018activation}
{Nwankpa et al.}, ``Activation functions: Comparison of trends in practice and research for deep learning,'' \emph{arXiv preprint arXiv:1811.03378}, 2018.

\bibitem{liew2016bounded}
S.~S. Liew, M.~Khalil-Hani, and R.~Bakhteri, ``Bounded activation functions for enhanced training stability of deep neural networks on visual pattern recognition problems,'' \emph{Neurocomputing}, vol. 216, pp. 718--734, 2016.

\bibitem{hoang2020ft}
L.-H. Hoang, M.~A. Hanif, and M.~Shafique, ``Ft-clipact: Resilience analysis of deep neural networks and improving their fault tolerance using clipped activation,'' in \emph{2020 Design, Automation \& Test in Europe Conference \& Exhibition (DATE)}.\hskip 1em plus 0.5em minus 0.4em\relax IEEE, 2020, pp. 1241--1246.

\bibitem{chen2020ranger}
Z.~Chen, G.~Li, and K.~Pattabiraman, ``A low-cost fault corrector for deep neural networks through range restriction,'' in \emph{2021 51st Annual IEEE/IFIP International Conference on Dependable Systems and Networks (DSN)}.\hskip 1em plus 0.5em minus 0.4em\relax IEEE, 2021, pp. 1--13.

\bibitem{9774635}
B.~Ghavami, M.~Sadati, Z.~Fang, and L.~Shannon, ``Fitact: Error resilient deep neural networks via fine-grained post-trainable activation functions,'' in \emph{2022 Design, Automation \& Test in Europe Conference \& Exhibition (DATE)}.\hskip 1em plus 0.5em minus 0.4em\relax IEEE, 2022, pp. 1239--1244.

\bibitem{8536419}
F.~F.~d. Santos, P.~F. Pimenta, C.~Lunardi, L.~Draghetti, L.~Carro, D.~Kaeli, and P.~Rech, ``Analyzing and increasing the reliability of convolutional neural networks on gpus,'' \emph{IEEE Transactions on Reliability}, vol.~68, no.~2, pp. 663--677, 2019.

\bibitem{9319148}
F.~Libano, P.~Rech, B.~Neuman, J.~Leavitt, M.~Wirthlin, and J.~Brunhaver, ``How reduced data precision and degree of parallelism impact the reliability of convolutional neural networks on fpgas,'' \emph{IEEE Transactions on Nuclear Science}, vol.~68, no.~5, pp. 865--872, 2021.

\bibitem{ruospo2021investigating}
A.~Ruospo, E.~Sanchez, M.~Traiola, I.~O’connor, and A.~Bosio, ``Investigating data representation for efficient and reliable convolutional neural networks,'' \emph{Microprocessors and Microsystems}, vol.~86, p. 104318, 2021.

\bibitem{9843614}
G.~Gambardella, N.~J. Fraser, U.~Zahid, G.~Furano, and M.~Blott, ``Accelerated radiation test on quantized neural networks trained with fault aware training,'' in \emph{2022 IEEE Aerospace Conference (AERO)}, 2022, pp. 1--7.

\bibitem{9897813}
N.~Cavagnero, F.~D. Santos, M.~Ciccone, G.~Averta, T.~Tommasi, and P.~Rech, ``Transient-fault-aware design and training to enhance dnns reliability with zero-overhead,'' in \emph{2022 IEEE 28th International Symposium on On-Line Testing and Robust System Design (IOLTS)}, 2022, pp. 1--7.

\bibitem{jacob2018quantization}
B.~Jacob, S.~Kligys, B.~Chen, M.~Zhu, M.~Tang, A.~Howard, H.~Adam, and D.~Kalenichenko, ``Quantization and training of neural networks for efficient integer-arithmetic-only inference,'' in \emph{Proceedings of the IEEE conference on computer vision and pattern recognition}, 2018, pp. 2704--2713.

\bibitem{reagen2016minerva}
{Reagen et al.}, ``Minerva: Enabling low-power, highly-accurate deep neural network accelerators,'' in \emph{ISCA}, 2016, pp. 267--278.

\bibitem{nguyen2019st}
{Nguyen et al.}, ``St-drc: Stretchable dram refresh controller with no parity-overhead error correction scheme for energy-efficient dnns,'' in \emph{DAC}, 2019, pp. 1--6.

\bibitem{yang2017bmxnet}
{Yang et al.}, ``Bmxnet: An open-source binary neural network implementation based on mxnet,'' in \emph{Proceedings of the 25th ACM international conference on Multimedia}, 2017, pp. 1209--1212.

\end{thebibliography}

\end{document}